\documentclass{article}
\usepackage{amssymb}

\usepackage[final]{corl_2025} 

\usepackage{hyperref}

\usepackage{url}
\usepackage{enumitem}
\usepackage{graphicx}
\usepackage{bm}
\usepackage{booktabs}
\usepackage{multirow}
\usepackage{colortbl}
\usepackage[dvipsnames]{xcolor}
\usepackage{wrapfig}
\usepackage{adjustbox}
\usepackage{subcaption}
\usepackage{amsmath}
\usepackage{pifont}
\usepackage{enumitem}
\let\oldding\ding
\renewcommand{\ding}[2][1]{\scalebox{#1}{\oldding{#2}}}

\usepackage{subcaption} 

\usepackage[table,xcdraw]{xcolor}
\usepackage{amssymb}
\usepackage{pifont}
\usepackage{makecell}  
\definecolor{myyellow}{RGB}{234,177,0} 
\definecolor{myblue}{RGB}{72,116,203} 
\usepackage{caption}
\title{ReasonPlan: Unified Scene Prediction and Decision Reasoning for Closed-loop Autonomous Driving}

\author{
  Xueyi Liu\textsuperscript{1,2,3} \and 
  Zuodong Zhong\textsuperscript{4} \and 
  Junli Wang\textsuperscript{1,2} \and
  Yuxin Guo\textsuperscript{1,2} \and
  Zhiguo Su\textsuperscript{3} \and
  Qichao Zhang\textsuperscript{1,2}\thanks{Corresponding author} \and
  Yun-Fu Liu\textsuperscript{3} \and
  Yinfeng Gao\textsuperscript{4} \and 
  Yupeng Zheng\textsuperscript{1,2} \and
  Qiao Lin\textsuperscript{3} \and
  Huiyong Chen\textsuperscript{3} \and
  Dongbin Zhao\textsuperscript{1,2}\color{Blue}\footnotemark[1] \and
  \vspace{0.1cm}
  \\
  \footnotesize
  \textsuperscript{1}SKL-MAIS, Institute of Automation, Chinese Academy of Sciences, Beijing, China \\
  \footnotesize
  \textsuperscript{2}School of Artificial Intelligence, University of Chinese Academy of Sciences, Beijing, China \\
  \footnotesize
  \textsuperscript{3}EACON, Fujian, China\\
  \footnotesize
  \textsuperscript{4}School of Automation and Electrical Engineering, University of Science and Technology Beijing, Beijing, China\\
}

\begin{document}
\maketitle

\begin{abstract}
Due to the powerful vision-language reasoning and generalization abilities, multimodal large language models (MLLMs) have garnered significant attention in the field of end-to-end (E2E) autonomous driving. However, their application to closed-loop systems remains underexplored, and current MLLM-based methods have not shown clear superiority to mainstream E2E imitation learning approaches. In this work, we propose ReasonPlan, a novel MLLM fine-tuning framework designed for closed-loop driving through holistic reasoning with a self-supervised Next Scene Prediction task and supervised Decision Chain-of-Thought process. This dual mechanism encourages the model to align visual representations with actionable driving context, while promoting interpretable and causally grounded decision making. We curate a planning-oriented decision reasoning dataset, namely PDR, comprising 210k diverse and high-quality samples. Our method outperforms the mainstream E2E imitation learning method by a large margin of $16.44\%$ L2 and $16.1$ driving score on Bench2Drive benchmark. Furthermore, ReasonPlan demonstrates strong zero-shot generalization on unseen DOS benchmark, highlighting its adaptability in handling zero-shot corner cases. Code and dataset will be found in \url{https://github.com/Liuxueyi/ReasonPlan}.

\end{abstract}



\section{Introduction}
Recently, end-to-end (E2E) autonomous driving presents a scalable, data-driven paradigm that has garnered increasing attention~\cite{hu2023planning, jiang2023vad,  yang2025uncad}. Despite its advantages in simplifying the driving pipeline, most existing E2E approaches rely on imitation learning~\cite{chen2024vadv2,sun2024sparsedrive} and exhibit limitations in complex, closed-loop environments. Specifically, they often suffer from causal confusion during interactive cases~\cite{de2019causal} and struggle to generalize to out-of-distribution scenarios~\cite{ross2011reduction}.
Recent progress in multimodal large language models (MLLMs)~\cite{openai2023gpt, li2024llava, wang2024qwen2} enables vision-language reasoning~\cite{lu2024deepseek} and zero-shot generalization~\cite{ chaiempowering} capabilities, offering new opportunities for E2E autonomous driving.


Recent efforts have explored dual-system frameworks~\cite{wang2023drivemlm, tian2024drivevlm, jiang2024senna}, LLM distillation for enhancing E2E driving~\cite{pan2024vlp, hegde2025distilling}, and direct trajectory prediction in textual form~\cite{shao2024lmdrive, tiantokenize, nie2024reason2drive}. While promising, these approaches predominantly operate in open-loop settings or exhibit suboptimal performance in closed-loop evaluations.
This limitation stems from their inability to perform context-aware reasoning and robust planning in closed-loop scenarios, where continuous adaptation to dynamic environments is essential~\cite{wendilu}.
We conclude three key challenges that limit the full exploitation of MLLMs' reasoning capabilities:
(1) \textbf{Insufficient utilization of visual information.} Existing methods~\cite{yang2024llm4drive} supervise the perception and decision-making process using only text or simply use predicted images as input for action generation~\cite{zheng2024doe}, leading to constrained scene comprehension and the accumulation of compounded errors.
(2) \textbf{No explicit reasoning process.} Prior works~\cite{shao2024lmdrive, tiantokenize} utilize multi-turn QAs fine-tuning LLMs to enhance the instruction-following ability, but they fall short of engaging the models’ Chain-of-Thought (CoT) reasoning capabilities.  
(3) \textbf{Lack of planning-oriented high-quality reasoning datasets.}
Consequently, the full potential of MLLMs for E2E closed-loop planning remains underexplored, and current MLLM-based methods have yet to demonstrate clear superiority over imitation learning techniques~\cite{hu2023planning, jiang2023vad, song2025don} in closed-loop benchmarks.

\begin{wrapfigure}{r}{0.4\linewidth}
\centering
\vspace{-0.4cm}
\includegraphics[width=5.5cm]{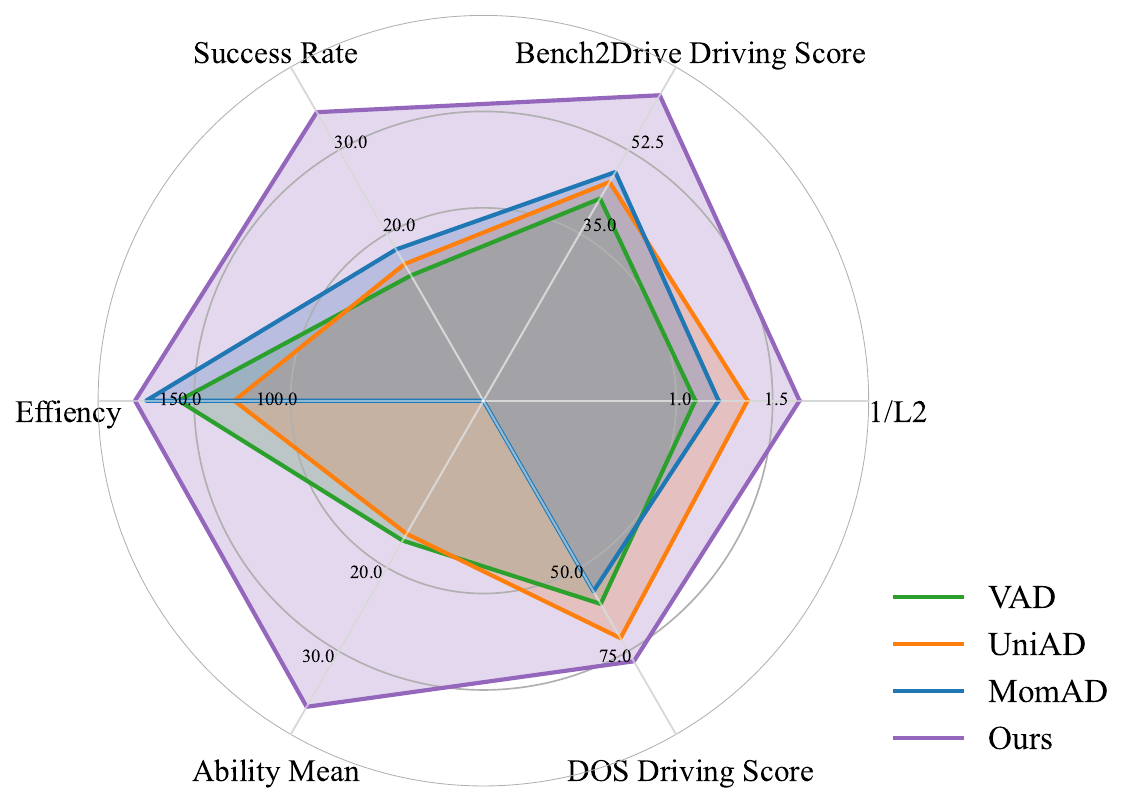}
\vspace{-0.2cm}
\caption{\small The proposed ReasonPlan achieves leading performance on most of metrics compared with E2E methods. }
\label{fig: holistic}
\vspace{-0.3cm}
\end{wrapfigure}
To address the challenges above, we explore both model architecture and training datasets. {From the model perspective}, we propose a novel MLLM fine-tuning framework, namely \textbf{ReasonPlan}, which effectively enhances the holistic reasoning capabilities in complicated closed-loop scenarios.
ReasonPlan comprises a \textbf{self-supervised Next Scene Prediction (NSP) task} and \textbf{supervised Decision Chain-of-Thought (DeCoT) process}. 
Specifically, motivated by recent advances in scene understanding and generative modeling~\cite{gao2024piwm, li2024enhancing}, we introduce a temporally NSP task that conditions on ego context to forecast future visual observations. This objective constrains image token representations in the latent space, enriching the model’s contextual understanding of driving scenarios.
To further leverage the reasoning and generalization capabilities under closed-loop and out-of-distribution evaluations, we incorporate explicit textual supervision over the DeCoT process. 

{From the dataset perspective}, we construct a large-scale instruction dataset tailored for closed-loop planning, called \textbf{PDR}, which contains 203,353 training samples and 11,047 testing samples. 
Using an automated annotation pipeline, PDR captures the complete decision reasoning process in training scenarios on the Bench2Drive~\cite{jia2024bench2drive}, including the following stages: {\color{myblue}\textit{Scene Understanding}}, {\color{myyellow}\textit{Traffic Sign Recognition}}, {\color{violet}\textit{Critical Object Identification for Risk Assessment}}, and {\color{orange}\textit{Meta Action}}. The dataset will be released publicly, serving as a foundation for learning structured and causally grounded decision reasoning.

ReasonPlan achieves a \textbf{driving score of $\textbf{64.01}$} and reduces the \textbf{L2 error by $\textbf{16.44\%}$} on Bench2Drive~\cite{jia2024bench2drive}, outperforming prior state-of-the-art E2E imitation learning models. Furthermore, it demonstrates \textbf{strong zero-shot generalization} on DOS~\cite{shao2023reasonnet}, highlighting its adaptability in decision-critical corner cases.

\textbf{Our key contributions are as follows. }(1) We propose ReasonPlan, a novel MLLM fine-tuning framework for complex closed-loop driving scenarios. The integration of NSP and DeCoT effectively couples the visual and language modalities, enabling a comprehensive decision reasoning process. (2) We construct PDR, a large-scale decision reasoning dataset via an automated annotation pipeline tailored for closed-loop planning, containing 210k diverse and high-quality samples. (3) As shown in Fig.~\ref{fig: holistic}, ReasonPlan demonstrates superior performance on the Bench2Drive under both open-loop and closed-loop settings, and shows strong zero-shot generalization on DOS scenarios.

    
    
    

\section{Related Work}
\label{sec:related work}
\textbf{End-to-End Autonomous Driving.}
Current mainstream E2E driving directly maps raw sensor inputs to trajectories based on imitation learning (IL).
Representative works such as UniAD~\cite{hu2023planning} and VAD~\cite{jiang2023vad} unify perception, prediction, and planning into a single framework, enabling joint optimization across the full pipeline.
To address planning uncertainty, SparseDrive~\cite{sun2024sparsedrive} adopts sparse representations in a multimodal planning framework, while UncAD~\cite{yang2025uncad} integrates uncertainty-aware online mapping.
VADv2~\cite{chen2024vadv2} models a probabilistic distribution over actions, enhancing robustness via action sampling.
However, these methods are evaluated in open-loop settings, where the models tend to overfit to specific ego-vehicle states~\cite{zhai2023ADMLP}.
DiffusionDrive~\cite{liao2024diffusiondrive} and GoalFlow~\cite{xing2025goalflow} explore a novel generative paradigm, leveraging diffusion models to predict diverse future trajectories in Navsim~\cite{dauner2024navsim}. Some other studies~\cite{wu2022trajectoryguided, song2025don} have adopted closed-loop evaluation in CARLA\cite{dosovitskiy2017carla} to assess driving robustness. However, those IL-based E2E still suffer from significant causal confusion~\cite{de2019causal} and limited generalization capabilities~\cite{ross2011reduction}.
To this end, we propose a MLLM-based E2E fine-tuning framework, aiming to harness their pretrained world knowledge and reasoning capabilities to address the challenges of closed-loop driving.

\textbf{Multimodal Large Language Models for Autonomous Driving.}
MLLMs have exhibited impressive capabilities in scene understanding and high-level reasoning across language, vision, and robotics domains~\cite{wei2022chain, shao2024visual, kimopenvla, tu2024online, chen2025conrft}, motivating their integration into autonomous driving systems.
Recent studies incorporate MLLMs into E2E frameworks via dual-system architectures and knowledge distillation~\cite{tian2024drivevlm, jiang2024senna, pan2024vlp, hegde2025distilling}.
DriveVLM~\cite{tian2024drivevlm} and Senna~\cite{jiang2024senna} leverage MLLMs to produce high-level driving intentions, which are subsequently refined by low-level policy modules for final trajectory generation.
VLP~\cite{pan2024vlp} and DiMA~\cite{hegde2025distilling} align key components of E2E systems with MLLMs, distilling abstract reasoning capabilities into lightweight planning heads.
Other methods~\cite{nie2024reason2drive, tiantokenize} adopt simple QAs fine-tuning to generate textual trajectories.
For example, Reason2Drive~\cite{nie2024reason2drive} enhances scene comprehension through CoT datasets and structured tokenization, while TOKEN~\cite{tiantokenize} improves long-tail planning by combining object-level perception with LLM-based reasoning.
Despite promising results, these approaches are limited to open-loop evaluations.
While some methods explore closed-loop settings~\cite{shao2024lmdrive, wang2023drivemlm, wendilu}, they often rely on simplified benchmarks such as Town05Long~\cite{prakash2021multi} or HighwayEnv~\cite{leurent2018environment}. LMDrive~\cite{shao2024lmdrive} introduces a language-based closed-loop framework for autonomous driving, lacking structured reasoning tasks.
SimLingo~\cite{renz2025simlingo}, built upon CarLLaVA~\cite{renz2024carllava}, introduces an action dreaming task to bridge language and control action spaces.
To fully exploit the reasoning potential of MLLMs in complex interactive scenarios, we propose a unified framework that tightly integrates visual and textual modalities, enabling comprehensive decision reasoning and zero-shot generalization in closed-loop scenarios.

\begin{figure}[tp!]
    \centering
    \includegraphics[width=1\textwidth]{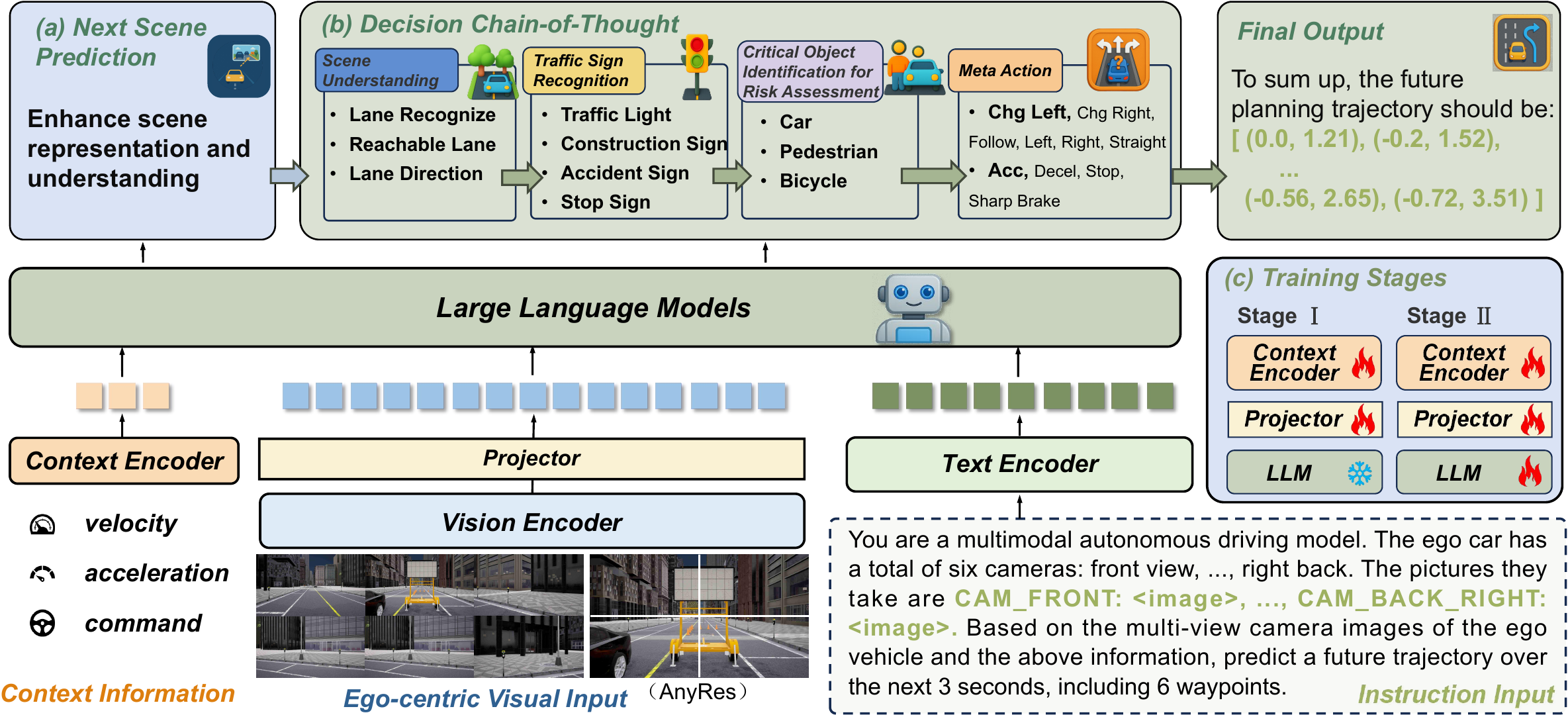}
    \caption{\small The pipeline of ReasonPlan, a holistic reasoning framework for closed-loop driving. It consists of two main modules: (a) the next scene prediction to enhance scene representation and understanding, which is conditioned on current context information; (b) the supervised decision CoT process to obtain the final planning trajectory. (c) the two training stages. }
    \label{fig: method}
    \vspace{-6mm}
\end{figure}

\section{Method}
\textbf{Overview.} 
The overall pipeline of ReasonPlan is illustrated in Fig.~\ref{fig: method}. Specifically, ReasonPlan comprises two components: (a) the self-supervised Next Scene Prediction task (Sec.~\ref{method: nsp}), which aims to enhance scene representation and understanding; (b) the supervised Decision Chain-of-Thought  process (Sec.~\ref{method: reason}) for reasoning and interpretable planning.  Moreover, (c) the framework is optimized in two stages (Sec.~\ref{method: stage}). 



\subsection{Self-supervised Next Scene Prediction (NSP)}
\label{method: nsp}
\begin{wrapfigure}{r}{0.4\linewidth}
\centering
\vspace{-4mm}
\includegraphics[width=6cm]{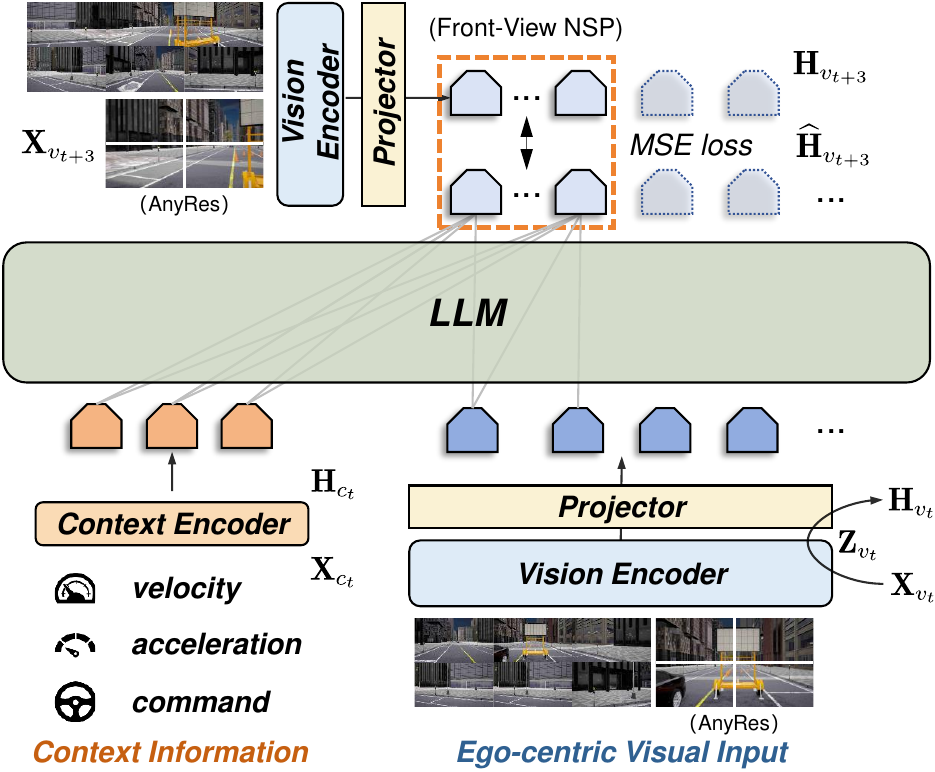}%
\vspace{-0.2cm}
\caption{\small The process of NSP task. }
\label{fig: NSP}
\end{wrapfigure}

To enhance scene understanding and effectively align the visual feature space with the semantic space of language, we introduce a temporally self-supervised NSP task, as illustrated in Fig.~\ref{fig: NSP}.
The input to our model consists of multi-view RGB images denoted as {$\mathbf{X}_{v_t} \in \mathbb{R}^{N \times W \times H \times 3}$},
where $N$ is the number of views, and $W$ and $H$ are the width and height of each image. 
We first resize the all-view images into grids.
Additionally, we adopt an AnyRes~\cite{liu2024llavanext} partitioning strategy that divides the front view $\mathbf{X}_{f_t}$ into four spatial grids to increase the model’s efficiency and ability to capture fine-grained spatial details.
Each gird is processed by the vision encoder SigLIP~\cite{zhai2023sigmoid}, yielding a visual feature tensor $\mathbf{Z}_{v_t} \in \mathbb{R}^{L_v \times D_v}$, where $L_v$ and $D_v$ denote the number of visual tokens per grid and the visual embedding dimension.
To align these features with the textual space, we apply a two-layer MLP projection module that maps $\mathbf{Z}_{v_t}$ to $\mathbf{H}_{v_t} \in \mathbb{R}^{L_v \times D_p}$, where $D_p$ denotes the language embedding dimension. This process can be formally written as: 
\begin{equation}
\mathbf{Z}_{v_t} = \texttt{SigLIP}(\texttt{AnyRes}(\mathbf{X}_{v_t})), \ \ \mathbf{H}_{v_t} = \texttt{MLP}(\mathbf{Z}_{v_t}),
\end{equation}

Considering the critical role of both vehicle dynamics and high-level driving intent, we introduce a context encoder, a two-layer MLP module, to embed the ego vehicle’s current velocity $\boldsymbol{v}$, acceleration $\boldsymbol{a}$, and driving command $\texttt{cmd}$ (e.g. turn left) into a context representation:
\begin{equation}
\mathbf{H}_{c_t} = \texttt{MLP}(\boldsymbol{v},\boldsymbol{a},\texttt{cmd}),
\end{equation}
This encoded context is then fused with the visual features $\mathbf{H}_{v_t}$ to guide the prediction of future scene representation. After the LLM, we can  estimate the latent visual embedding $\mathbf{\hat{H}}_{v_{t+3}}$. 
For self-supervised labeling, we utilize the multi-view images at a 3-second time horizon $\mathbf{{X}}_{v_{t+3}}$ as the target future frame. These images are processed through the identical SigLIP  encoder and MLP projection after AnyRes grid processing, yielding latent visual representations $\mathbf{{H}}_{v_{t+3}}$ as self-supervision signals. 
In autonomous driving, the front-view image contains the most semantically informative content, capturing critical cues for trajectory planning. Our ablation study (in Appendix~\ref{sup: ablation}) also reveals that, for NSP task, training front-view achieves performance on par with the full-view setting. To improve training efficiency and eliminate redundant computation, we uniformly adopt the front-view NSP in subsequent experiments. While the MSE loss is computed solely on the front-view, the full image features are retained as condition, providing comprehensive contextual information for the subsequent reasoning process.
\begin{equation}
\mathbf{\hat{H}}_{v_{t+3}} = \texttt{LLM}(\texttt{Concat}(\mathbf{H}_{c_t}, \mathbf{H}_{v_t})), \ \  \mathcal{L}_\text{image}=\Vert \mathbf{\hat{H}}_{v_{t+3}}{[: \texttt{front}]}-\mathbf{H}_{v_{t+3}}{[: \texttt{front}]}\Vert^2,
\end{equation}

We employ a Mean Squared Error (MSE) loss in latent feature space to enforce consistency between predicted and future visual representations.
By conditioning on both dynamic and contextual cues, our model anticipates future perceptual states aligned with intended maneuvers. Through a temporally self-supervised NSP task, it integrates ego motion and scene context to enhance the LLM's spatial awareness, enabling comprehensive scene understanding and context-aware forecasting. 

\subsection{Decision Chain-of-Thought (DeCoT)}
\label{method: reason}
Following the NSP task, the DeCoT process in the textual space is pivotal in our framework. ReasonPlan leverages the pretrained knowledge of LLM, integrating visual tokens and textual instructions to generate interpretable decision-making processes and produce executable trajectories in text form.

As shown in Fig.~\ref{fig: method}, the system prompt and navigation instructions $\mathbf{X}_p$ are first tokenized into token IDs and then encoded into textual embeddings $\mathbf{H}_p^{L_p \times D_p}$ using a text encoder, where $L_p$ denotes the number of text tokens. 
To enable effective multimodal alignment, we introduce a special \texttt{<image>} token into the tokenizer vocabulary. To preserve spatial awareness and allow the model to distinguish among different viewpoints, we explicitly annotate each token with its corresponding camera perspective in the prompt (e.g., \texttt{CAM\_FRONT: <image>}, ..., \texttt{CAM\_BACK: <image>}). 
These \texttt{<image>} tokens are dynamically substituted with the corresponding visual embeddings processed by the image encoder, enabling seamless integration of textual and visual modalities.
To better support the implementation of NSP and DeCoT, we introduce six additional special tokens beyond \texttt{<image>}. The input and target sequences for the LLM are structured as follows:
\[
\small{\texttt{User: \{velocity\} \{acceleration\} \{navigation command\} \{image tokens\}\_t \{prompt\}}}.
\]
\[
\small{\texttt{Assistant: [BOS] [BOI] \{image tokens\}\_t+3 [EOI] [BOT] \{reasoning steps\} [EOT] }}
\]
\[
\small{\texttt{\{generated trajectory\} [EOS]}}.
\]
Where [BOS] and [EOS] are the original special tokens in the text tokenizer, [BOI] and [EOI] marking the
start and end of the image tokens. Similarly, [BOT] and [EOT] represent the begin and end of reasoning process.

To perform human-like reasoning, we introduce explicit supervision over the language model's intermediate reasoning steps, thereby enhancing its ability to handle complex decision-making tasks.  Leveraging both visual and textual modalities, ReasonPlan performs a planning-oriented thinking process before the final planning, including {\color{myblue}\textit{Scene Understanding}}, {\color{myyellow}\textit{Traffic Sign Recognition}}, {\color{violet}\textit{Critical Object Identification for Risk Assessment}}, and {\color{orange}\textit{Meta Action}}. 
For a sequence of length $L$, we compute the probability of the target answers $\mathbf{X}_{{a}}$ and the Cross-Entropy (CE) loss:
\begin{equation}
p(\mathbf{X}_{a}|\mathbf{X}_{{v}},\mathbf{X}_{{p}})=\prod_{i=1}^L p(\boldsymbol{x}_i|\mathbf{X}_{{v}},\mathbf{X}_{{p},<i},\mathbf{X}_{{a},<i}),
\mathcal{L}_{\text{text}} = -\log p(\mathbf{X}_{a}|\mathbf{X}_{{v}},\mathbf{X}_{{p}}),
\end{equation}
where $\mathbf{X}_{{p}, <i}$ and $\mathbf{X}_{{a}, <i}$ are the instruction and answer tokens in all turns before the current prediction token $\boldsymbol{x}_i$, respectively.

Unlike traditional multi-turn QAs paradigms, ReasonPlan performs multi-step reasoning in a single forward pass, effectively leveraging the commonsense reasoning capabilities of LLM.

\subsection{Training Strategy}
\label{method: stage}


The final training objective is formulated as a weighted combination of the self-supervised image prediction loss and the language reasoning loss:
\begin{equation}
\mathcal{L}_{\text{total}} = \lambda_1 \cdot \mathcal{L}_{\text{image}} + \lambda_2 \cdot \mathcal{L}_{\text{text}}
\end{equation}
where $\lambda_1$ and $\lambda_2$ are the weights for the visual-space MSE loss and the textual-space CE loss, respectively.
The overall framework is optimized in two stages, as shown in Fig.~\ref{fig: method} (c). In the first stage, we train the projection module and context encoder using non-decision supervision to align the visual feature space with the textual semantic space, while extracting contextual cues from ego vehicle states and navigation commands. In the second stage, we employ the collected PDM dataset to jointly fine-tune the projection module, context encoder, and the LLM backbone, transferring pretrained commonsense reasoning capabilities to complex  driving scenarios.

\section{PDR: Planning-oriented Decision Reasoning Dataset}
While various QA datasets have been introduced for autonomous driving~\cite{wang2023drivemlm, tiantokenize, nie2024reason2drive}, they are primarily designed for scene understanding and open-loop evaluation, and thus fail to assess actual driving performance in closed-loop settings. 
To bridge this gap, we construct a large-scale, high-quality decision reasoning dataset, namely PDR, which focuses on trajectory planning and comprises 210k diverse and high-quality samples. 
We develop an automated annotation pipeline tailored for complex decision-making in closed-loop scenarios. PDR intends to leverage the LLM's reasoning and generalization capabilities in dynamic driving environments.

\begin{figure}[t]
    \centering
    \includegraphics[width=0.75\textwidth]{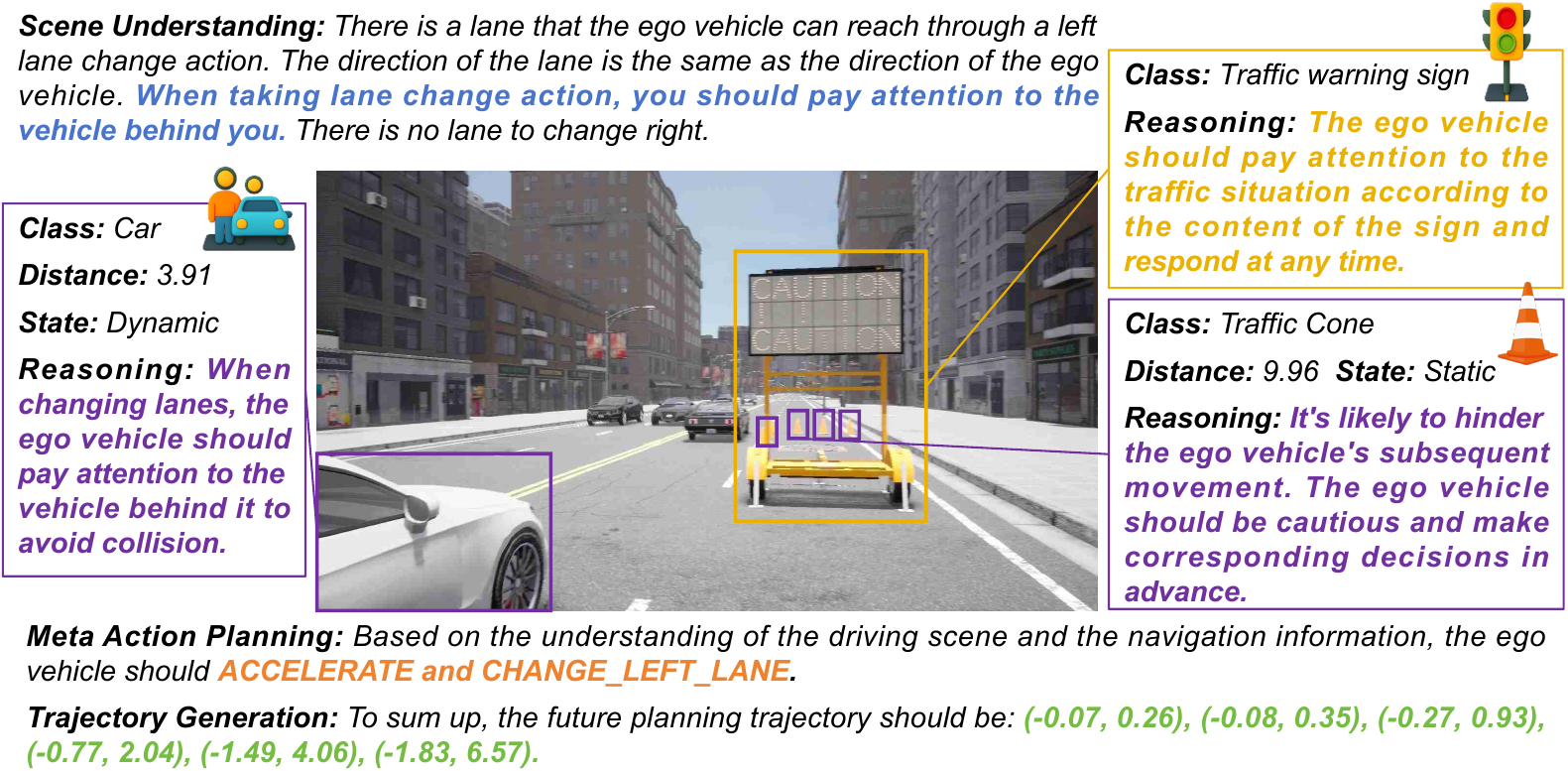}
    \caption{\small An annotated sample of the PDR dataset.}
    \label{fig: data}
    \vspace{-0.4cm}
\end{figure}

To construct a reliable reasoning dataset, we extend the ground-truth annotations provided by Bench2Drive~\cite{jia2024bench2drive} with structured reasoning labels. The reasoning traces have undergone thorough human verification to ensure consistency and interpretability across diverse scenarios.
As shown in Fig.~\ref{fig: data}, the reasoning process comprises the following stages: {\color{myblue}\textit{Scene Understanding}}, {\color{myyellow}\textit{Traffic Sign Recognition}}, {\color{violet}\textit{Critical Object Identification for Risk Assessment}}, and {\color{orange}\textit{Meta Action}}. In Appendix~\ref{sup: dataset}, we detail the statistics of PDR and the annotation pipeline for each component.

\section{Experiments}
\subsection{Benchmark and Evaluation Metrics}
\textbf{Benchmark.} 
We evaluate the closed-loop driving performance of ReasonPlan on Bench2Drive~\cite{jia2024bench2drive}, which features challenging interactive scenarios based on the Carla leaderboard v2. To further assess its reasoning capabilities, we also conduct zero-shot evaluations on DriveOcclusionSim (DOS)~\cite{shao2023reasonnet}, a suite of complex scenarios requiring models to infer global context from dynamic driving environments.
For ablation studies, we use the Dev10~\cite{jia2024bench2drive} for quick validation.

\textbf{Implementation details. }
Our framework processes six surround-view images captured at an original resolution 1600×900, which are resized and encoded using the AnyRes~\cite{liu2024llavanext} 
strategy to produce ten spatial grids of size 384×384. Our framework employs SigLIP~\cite{zhai2023sigmoid} as the vision encoder.
For the language model, we adopt Qwen-0.5B~\cite{wang2024qwen2}, a lightweight yet capable LLM that balances efficiency and reasoning capability. 
The learning rate is fixed at 5e-5, and the weights for both $\mathcal{L}_{\text{image}}$ and $\mathcal{L}_{\text{text}}$ are set to 1.0. 
More details in Appendix~\ref{sup: implementation}.

\textbf{Evaluation Metrics.} For open-loop, we report the L2 distance between 2s predicted trajectories and expert trajectories. For closed-loop, we adopt the metrics from \cite{jia2024bench2drive}: (1) Route Completion (RC): percentage of completed route;
(2) Infraction Score (IS): traffic violation penalty score;
(3) Driving Score (DS): RC × IS (overall performance metric);
(4) Success Rate (SR): percentage of infraction-free, timely completed episodes;
(5) Efficiency (Effi): ego speed relative to neighboring vehicles' average;
(6) Comfort (Comf): compliance with motion smoothness thresholds. 

\begin{table*}[t]
\centering
\small
\caption{\small \textbf{Planning and Multi-Ability Performance in Bench2Drive.} * denotes expert feature distillation. M=Merging. O=Overtaking. EB=Emergency Brake. Give Way=GW. TS=Traffic Sign.}
\label{tab:unified}
\resizebox{\linewidth}{!}{
\begin{tabular}{l|c|cccc|ccccc|c}
\toprule
\multirow{2}{*}{\textbf{Method}} & \textbf{Open-loop} & \multicolumn{4}{c|}{\textbf{Closed-loop Metric}} & \multicolumn{5}{c|}{\textbf{Ability} (\%) $\uparrow$} & \textbf{Ability} \\ \cmidrule{2-11}
& Avg. L2 $\downarrow$ & \cellcolor{gray!15} DS $\uparrow$ & SR $\uparrow$ & Effi $\uparrow$ & Comf $\uparrow$ & M & O & EB & GW & TS & Mean \\
\midrule
TCP*~\citep{wu2022trajectoryguided} & 1.70 & \cellcolor{gray!15} 40.70 & 15.00 & 54.26 & 47.80 & 17.50 & 13.63 & 20.00 & 10.00 & 6.81 & \cellcolor{gray!15}13.59 \\
ThinkTwice*~\citep{jia2023thinktwice} & 0.95 & \cellcolor{gray!15} 62.44 & 31.23 & 69.33 & 16.22 & 13.72 & 22.93 & 52.99 & 50.00 & 47.78 & \cellcolor{gray!15}37.48 \\
DriveAdapter*~\citep{jia2023driveadapter} & 1.01 & \cellcolor{gray!15} \textbf{64.22} & \textbf{33.08} & 70.22 & 16.01 & 14.55 & 22.61 & 54.04 & 50.00 & 50.45 & \cellcolor{gray!15}\textbf{38.33} \\
\midrule
AD-MLP~\citep{zhai2023ADMLP} & 3.64 & \cellcolor{gray!15} 18.05 & 0.00 & 48.45 & 22.63 & 0.00 & 0.00 & 0.00 & 0.00 & 0.00 & \cellcolor{gray!15}0.00 \\
UniAD-Tiny~\citep{hu2023planning} & 0.80 & \cellcolor{gray!15} 40.73 & 13.18 & 123.92 & 47.04 & 7.04 & 10.00 & 21.82 & 20.00 & 14.61 & \cellcolor{gray!15}14.69 \\
UniAD-Base~\citep{hu2023planning} & 0.73 & \cellcolor{gray!15} 45.81 & 16.36 & 129.21 & 43.58 & 12.16 & 20.00 & 23.64 & 10.00 & 13.89 & \cellcolor{gray!15}15.94 \\
VAD~\citep{jiang2023vad} & 0.91 & \cellcolor{gray!15} 42.35 & 15.00 & 157.94 & 46.01 & 7.14 & 20.00 & 16.36 & 20.00 & 20.22 & \cellcolor{gray!15}16.75 \\
MomAD~\cite{song2025don} & 0.82 & \cellcolor{gray!15} 47.91 & 18.11 & 174.91 & \textbf{51.20} & - & - & - & - & - & - \\
\midrule
\rowcolor{orange!20}
ReasonPlan (\textbf{Ours}) & \textbf{0.61} & \textbf{64.01} & \textbf{34.55} & \textbf{180.64} & 25.63 & \textbf{37.50} & \textbf{26.67} & \textbf{33.30} & \textbf{40.00} & \textbf{45.79} & \textbf{36.66} \\
\bottomrule
\end{tabular}}
\vspace{-4mm}
\end{table*}

\subsection{Comparison with State-of-the-arts}
\textbf{SOTA Performance in both Open-Loop and Closed-Loop Evaluations. }
As reported in Tab.~\ref{tab:unified}, \textbf{ReasonPlan} achieves the best open-loop performance with \textbf{0.61} L2 error, demonstrating superior trajectory prediction accuracy. 
In closed-loop evaluation, ReasonPlan delivers competitive performance compared to SOTA method DriveAdapter~\cite{jia2023driveadapter} 
which utilizes privileged expert feature distillation.
Specifically, it achieves a DS of \textbf{64.01}, significantly outperforming the non-distillation IL-based SOTA method, MomAD~\cite{song2025don} by \textbf{16.1}. Furthermore, ReasonPlan improves the success rate by \textbf{16.44\%} compared to MomAD~\cite{song2025don}.
ReasonPlan also achieves the \textbf{highest efficiency score of 180.64} among all evaluated methods, demonstrating an effective and proactive driving policy. The comfort score of 25.63 reflects a common trade-off between trajectory agility and smoothness. Nonetheless, the comfort margin could be further optimized via post-smoothing or low-level controller tuning.
Besides, \textbf{ReasonPlan} achieves a \textbf{Mean Ability} score of \textbf{36.66\%} across diverse driving scenarios, significantly outperforming all E2E baselines. These results underscore the model’s strong capability to reason across diverse driving intents and validate its robustness under complex, multi-intent scenarios.

\textbf{Strong Zero-Shot Generalization in Complex Reasoning Scenarios.} To evaluate the out-of-distribution generalization capability of \textbf{ReasonPlan}, we conduct zero-shot closed-loop evaluations on the \textbf{DOS} benchmark, as summarized in Tab.~\ref{tab:dos}. Notably, no methods are trained on DOS, ensuring a purely zero-shot setting. 
ReasonPlan demonstrates consistent and superior performance under these conditions, achieving the highest average DS of $\textbf{78.02}$. 
These results highlight the strong generalization capacity of ReasonPlan, driven by its holistic reasoning pipeline, which enables robust and safe decision-making even in unseen scenarios.

\begin{table}[t]
\centering
\small
\caption{\small \textbf{Planning Performance in DOS}. DOS includes four types of challenging occlusion-reasoning driving scenarios. DOS\_01 = Parked Cars, DOS\_02 = Sudden Brake, DOS\_03 = Left Turn, DOS\_04 = Red Light Infraction. The details about each scenario can be found in Appendix~\ref{sup: benchmark}.}
\label{tab:dos}
\renewcommand{\arraystretch}{1.2}
\setlength{\tabcolsep}{4pt}
\resizebox{\linewidth}{!}{
\begin{tabular}{l|ccc|ccc|ccc|ccc|c}
\toprule
\multirow{2}{*}{\centering \textbf{Method}} & \multicolumn{3}{c|}{\textbf{DOS\_01}} & \multicolumn{3}{c|}{\textbf{DOS\_02}} & \multicolumn{3}{c|}{\textbf{DOS\_03}} & \multicolumn{3}{c|}{\textbf{DOS\_04}} & \multirow{2}{*}{\textbf{Average$\uparrow$}} \\
\cmidrule{2-13}
& RC$\uparrow$ & IS$\uparrow$ & DS$\uparrow$ & RC$\uparrow$ & IS$\uparrow$ & DS$\uparrow$ & RC$\uparrow$ & IS$\uparrow$ & DS$\uparrow$ & RC$\uparrow$ & IS $\uparrow$& DS$\uparrow$ & \\
\midrule
UniAD-Tiny~\citep{hu2023planning} & 85.98 & 0.58 & \cellcolor{gray!15}49.95 & 90.91 & 0.78 & \cellcolor{gray!15}69.74 & 50.01 & \textbf{0.97} & \cellcolor{gray!15}49.83 & 89.25 & 0.687 &\cellcolor{gray!15} 62.18 & \cellcolor{gray!15}57.93 \\
UniAD-Base~\citep{hu2023planning} & 95.65 & 0.69 & \cellcolor{gray!15}66.00 & 95.46 & 0.73 & \cellcolor{gray!15}69.32 & 93.78 & 0.78 & \cellcolor{gray!15}73.18 & 79.07 & 0.97 & \cellcolor{gray!15}75.87 & \cellcolor{gray!15}71.09 \\
VAD~\citep{jiang2023vad} & 86.97 & 0.71 & \cellcolor{gray!15}58.46 & 78.80 & 0.78 & \cellcolor{gray!15}57.84 & 74.75 & 0.85 & \cellcolor{gray!15}61.16 & 82.91 & 0.82 & \cellcolor{gray!15}65.88 & \cellcolor{gray!15}60.84 \\
MomAD~\citep{song2025don} & 64.86 & \textbf{0.86} & \cellcolor{gray!15}53.90 & 82.71 & 0.80 & \cellcolor{gray!15}63.36 & 64.87 & 0.91 & \cellcolor{gray!15}56.00 & 55.06 &\textbf{ 1.00} & \cellcolor{gray!15}55.06 & \cellcolor{gray!15}57.08 \\
LMDrive~\citep{shao2024lmdrive} & \textbf{100.00} & 0.64 & \cellcolor{gray!15}64.00 & \textbf{100.00} & 0.62 & \cellcolor{gray!15}62.00 & 94.42 & 0.78 & \cellcolor{gray!15}73.65 & \textbf{91.16} & 0.64 & \cellcolor{gray!15}58.48 & \cellcolor{gray!15}64.53 \\
\rowcolor{orange!20}
ReasonPlan (\textbf{Ours}) & 99.78 & 0.68 & \textbf{67.77}& 99.65 & \textbf{0.92} &\textbf{91.65} &\textbf{ 95.53} & 0.84 & \textbf{80.57} & 90.65 & 0.87 & \textbf{77.85} & \textbf{78.02} \\
\bottomrule
\end{tabular}}
\end{table}

\begin{figure}[tp!]
    \centering
    \includegraphics[width=1\textwidth]{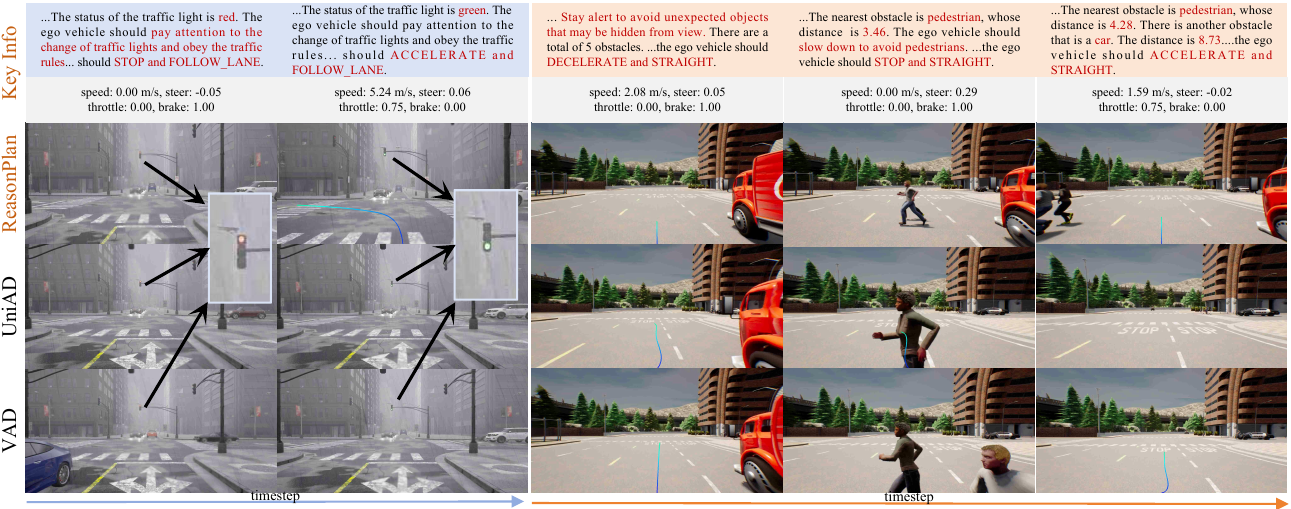}
    \caption{\small Qualitative comparison of ReasonPlan with baselines. The left case is the signalized junction within Bench2Drive. While baseline methods stall at green lights due to misinterpreting signal changes, ReasonPlan accurately detects the transition and proceeds safely through intersections. The right case is the pedestrian emerging scenario within DOS. While other methods fail to react in time and result in a collision, ReasonPlan anticipates the risk by decelerating early and executing a timely stop upon detection.}
    \label{fig: case}
    \vspace{-4mm}
\end{figure}

\subsection{Qualitative Results}
Fig.~\ref{fig: case} illustrates qualitative results of ReasonPlan in two representative closed-loop evaluation scenarios. The figures showcase the DeCoT reasoning process and the corresponding predicted trajectories. Compared to baseline methods, ReasonPlan exhibits superior performance in navigating complex intersections and handling unseen scenarios. 

\subsection{Ablation Studies and Analysis}
In this section, through detailed ablation studies, we  validate the effectiveness of our proposed methods and datasets through model and reasoning steps ablations.

\textbf{NSP effectively models dynamic scene transitions and enhances spatial planning. }
The NSP module introduces fine-grained visual understanding by applying dense supervision on image tokens. Through the temporally predictive task, it facilitates improved 3D spatial reasoning and enhances downstream planning performance (Tab.~\ref{subtab:ablation-model} ID 2). However, without explicit reasoning over decision-making, NSP alone fails to ensure traffic compliance, resulting in lower IS.

\textbf{DeCoT consistently facilitates planning through  structured reasoning.} DeCoT decomposes complicated decision-making into interpretable reasoning steps with direct supervision, thereby improving the model’s capacity to handle intricate scenarios (Tab.~\ref{subtab:ablation-model} ID 3).

\textbf{NSP and DeCoT are complementary and synergistic.} Integrating both NSP and DeCoT achieves the best overall performance (Tab.~\ref{subtab:ablation-model} ID 4), as NSP provides rich visual context for planning, while DeCoT enforces structured reasoning to regulate driving behavior.  These components empower ReasonPlan to perform unified, interpretable, and effective E2E planning in complex scenarios.


\newcommand{\AblationModel}
{{\footnotesize
\begin{tabular}{c|c|c|ccc}
\toprule
\textbf{ID} & NSP & DeCoT & RC$\uparrow$ & IS$\uparrow$ & DS$\uparrow$ \\
\midrule
1 & \color{red}\ding{55} & \color{red}\ding{55} & 58.79 & \textbf{0.79} & 41.84 \\
2 & \color{green}\ding{51} & \color{red}\ding{55} & \textbf{95.03} & 0.54 & 52.61 \\
3 & \color{red}\ding{55} & \color{green}\ding{51} & 73.73 & 0.76 & 53.97 \\
4 & \color{green}\ding{51} & \color{green}\ding{51} & 89.29 & 0.65 & \textbf{57.83} \\
\bottomrule
\end{tabular}}}

\newcommand{\AblationData}
{{\footnotesize
\begin{tabular}{c|c|c|c|c|ccc}
\toprule
\textbf{ID} & {\color{myblue}SU} & {\color{myyellow}TS} & {\color{violet}CO} & {\color{orange}MA} & RC$\uparrow$ & IS$\uparrow$ & DS$\uparrow$ \\
\midrule
1  & \color{red}\ding{55} & \color{green}\ding{51} & \color{green}\ding{51} & \color{green}\ding{51} & 69.02 & 0.65 & 43.28 \\
2  & \color{green}\ding{51} & \color{red}\ding{55} & \color{green}\ding{51} & \color{green}\ding{51} & 74.94 & 0.59 & 45.65 \\
3  & \color{green}\ding{51} & \color{green}\ding{51} & \color{red}\ding{55} & \color{green}\ding{51} & 76.54 & 0.55 & 43.98 \\
4  & \color{green}\ding{51} & \color{green}\ding{51} & \color{green}\ding{51} & \color{red}\ding{55} & 64.25 & \textbf{0.73} & 42.61 \\
5  & \color{green}\ding{51} & \color{green}\ding{51} & \color{green}\ding{51} & \color{green}\ding{51} & \textbf{89.29} & 0.65 & \textbf{57.83} \\
\bottomrule
\end{tabular}}}

\begin{table}[!ht]
\setlength\tabcolsep{3pt}
    \centering
    \caption{\small Ablations on (a) Each Module and (b) Reasoning Steps (Dev10). NSP=Next Scene Prediction. DeCoT=Decision Chain-of Thought. {\color{myblue}SU=Scene Understanding}, 
{\color{myyellow}TS=Traffic Sign Recognition}, 
{\color{violet}CO=Critical Object Identification for Risk Assessment}, 
{\color{orange}MA=Meta Action}. }
    \begin{subtable}{0.5\linewidth}
        \centering
        \renewcommand{\arraystretch}{1}
        \AblationModel
        \caption{\small Model architecture designs.}
        \label{subtab:ablation-model}
    \end{subtable}\hfill
    \begin{subtable}{0.5\linewidth}
        \centering
        \renewcommand{\arraystretch}{1}
        \AblationData
        \caption{\small Individual Reasoning Steps. }
        \label{subtab:ablation-data}
    \end{subtable}
    \label{tab:two-ablation}
    \vspace{-6mm}
\end{table}

\textbf{Structured and complete reasoning steps lead to the best driving performance.}
As shown in Tab.~\ref{subtab:ablation-data}, removing any single component from the full reasoning pipeline degrades the model’s ability to handle complex scenarios.
Specifically, omitting the {\color{orange}\textit{Meta Action}} step, which represents the driving decision, leads to a significant drop in DS.
These results emphasize the importance of fine-grained reasoning for safe and robust decision-making. Moreover, these findings underscore the quality and effectiveness of the PDR dataset generated by our automated annotation pipeline.



\section{Conclusion}
\label{sec:conclusion}
In this work, we present ReasonPlan, a novel fine-tuning framework that adapts MLLMs for complex closed-loop scenarios. ReasonPlan introduces a self-supervised Next Scene Prediction task and explicit Decision Chain-of-Thought process, enabling unified integration of visual and textual modalities for interpretable planning.
Through comprehensive evaluations on Bench2Drive and DOS, ReasonPlan achieves excellent performance in both open-loop and closed-loop settings. Notably, it demonstrates strong zero-shot generalization on unseen tasks, underscoring its robustness and potential for real-world deployment.
Our results highlight the promise of MLLM-based frameworks in bridging high-level reasoning and low-level planning, paving the way for more cognitive and generalizable autonomous driving systems.


\section{Limitations}
Firstly, while ReasonPlan leverages strong reasoning capabilities to address challenges in complex reasoning and out-of-distribution scenarios, its reliance on MLLMs introduces non-negligible inference latency. Although the LLM of 0.5B size is currently acceptable on real-time deployment, larger sizes such as 7B are still difficult to deploy in real time. Nonetheless, emerging techniques in latent-space reasoning offer promising directions to improve both the efficiency and responsiveness of the framework. In addition, can large reasoning models such as o1 or DeepSeek-R1 provide better reasoning capabilities for closed-loop autonomous driving is worth further analysis.
Secondly, ReasonPlan employs a holistic reasoning framework where actions are represented as unimodal textual outputs. A promising future direction is to decouple reasoning and action generation—delegating decision-making solely to the LLM, while using a specialized generative model to synthesize multimodal trajectories. This modular design, inspired by recent successes in robotics, may further enhance flexibility and scalability.
Thirdly, similar to most end-to-end frameworks, ReasonPlan relies on supervised fine-tuning over offline datasets, which limits its ability to learn from interactive feedback and may lead to occasional collisions (see Appendix~\ref{sup: case}). Post-training with reinforcement learning or integrating environment-aware adaptation mechanisms might be a future direction to address this limitation.
Last but not least, it remains unclear whether MLLMs are the most suitable foundation model for end-to-end autonomous driving and can well align visual language and actions. The VLA foundation model for autonomous driving should be further investigated.
\section{Acknowledgement}
{This work was supported by the National Key Re-search and Development Program of China under Grant 2022YFA1004000, in part by Beijing Natural Science Foundation under Grant L253007 and 4242052 , in part by the State Key Laboratory of Safety Intelligent Mining in Non-coal Open-pit Mines, National Mine Safety Administration (Grant No.2024-ZD04).}
\clearpage


\bibliography{example}  

\clearpage
\setcounter{page}{13}
\setcounter{figure}{0}
\setcounter{table}{0}
\renewcommand\thefigure{A\arabic{figure}} 
\renewcommand\thetable{A\arabic{table}}
\newcommand{\maketitlesupplementary}{
  \begin{center}
    {\LARGE \textbf{ReasonPlan: Unified Scene Prediction and Decision}}\\[1.5ex]
    {\LARGE \textbf{Reasoning for Closed-loop Autonomous Driving}}\\[2.5ex]
    {\large Supplementary Material}\\[2ex]
  \end{center}
  \vspace{2ex}
}

\maketitlesupplementary
\appendix

\noindent We provide supplementary material to complement the main paper, arranged as follows:
\begin{itemize}
    \item Appendix~\ref{sup: dataset}: Details on the PDR dataset.
    \item Appendix~\ref{sup: benchmark}: Benchmarks.
    \item Appendix~\ref{sup: implementation}: Implementation Details.
    \item Appendix~\ref{sup: ablation}: More Ablation Experiments.
    \item Appendix~\ref{sup: case}: More Case Study.
\end{itemize}
 
\section{Details on the PDR dataset}
\label{sup: dataset}
\subsection{Annotation Procedure}
Our dataset is sourced from Bench2Drive~\cite{jia2024bench2drive}, an offline dataset collected in highly challenging scenarios from Carla Leaderboard v2. 
Bench2Drive leverages the CARLA simulator to collect primitive per-frame annotations, including ego-vehicle states, sensor data, and detailed metadata on surrounding vehicles, pedestrians, and traffic signs. This enables a fully automated annotation process, eliminating the potential subjectivity and labor-intensiveness of manual labeling. Furthermore, it provides accurate information such as the relative distances between the ego vehicle and surrounding entities, significantly enhancing the dataset’s accuracy and reliability for reasoning tasks.
Specifically, to ensure high-quality reasoning data, the entire annotation process consists of the following stages:

\textbf{{\color{myblue}\textit{Scene Understanding}}.} 
In autonomous driving, the ego vehicle must continuously perceive and reason about its surroundings to enable safe and efficient trajectory planning. Leveraging high-definition (HD) map information, our system accurately identifies the driving context—such as highways, intersections, and ramp entrances—by parsing road topology and lane semantics, including lane count, types, and connectivity. Based on the recognized scenario, the system delineates the ego vehicle’s drivable area and enumerates all feasible candidate routes. These include adjacent lanes for potential lane changes, upcoming forks, and reachable target lanes. Such structured scene understanding serves as a critical foundation for downstream decision-making and motion planning, enabling the formulation of informed lane-changing and turning strategies.

\textbf{{\color{myyellow}\textit{Traffic Sign Recognition}}.}
Our dataset includes comprehensive annotations for diverse traffic signage, encompassing traffic signals, warning signs, and construction indicators. Warning and construction signs are particularly essential, as they offer prior knowledge about potential hazards such as congestion or partial lane closures. Upon detection, the system anticipates disruptions and adapts its plan through maneuvers such as preemptive lane changes. For traffic lights, we explicitly annotate three discrete signal states—red, green, and yellow—allowing for fine-grained decision logic in regulated environments.

\textbf{{\color{violet}\textit{Critical Object Identification for Risk Assessment}}.}
Robust identification of critical dynamic objects and the assessment of their associated risk are fundamental to safe motion planning. We annotate all potentially influential obstacles within each scene, filtering candidates based on interactive lane semantics and proximity to the ego vehicle. Using ground-truth 3D annotations, we extract fine-grained object attributes, including category, position, and motion state. These features are further used to reason about interaction dynamics—such as whether an object may trigger acceleration, deceleration, lane changes, or emergency maneuvers. The resulting risk estimations are translated into textual representations, facilitating downstream planning modules grounded in language-based reasoning.

\textbf{{\color{orange}\textit{Meta Action}}.}
To emulate high-level human driving intent, we annotate expert trajectories with structured meta-actions across lateral and longitudinal dimensions. Lateral decisions include six motion primitives: lane follow, left/right lane change, left/right turn, and straight, capturing diverse driving intents across typical scenarios. For longitudinal behavior, we annotate acceleration, deceleration, stop, and emergency braking to reflect varying control responses. Lateral labels are derived from observed deviations in trajectory geometry, while longitudinal annotations follow expert speed profiles. Given the noise inherent in reinforcement learning (RL)-based expert policies, we apply a low-pass filter to suppress high-frequency fluctuations and enhance annotation stability. Finally, these meta actions are serialized into language-based formats to support integration with text-conditioned planning modules.

\begin{figure}[tp!]
    \centering
    \includegraphics[width=0.8\textwidth]{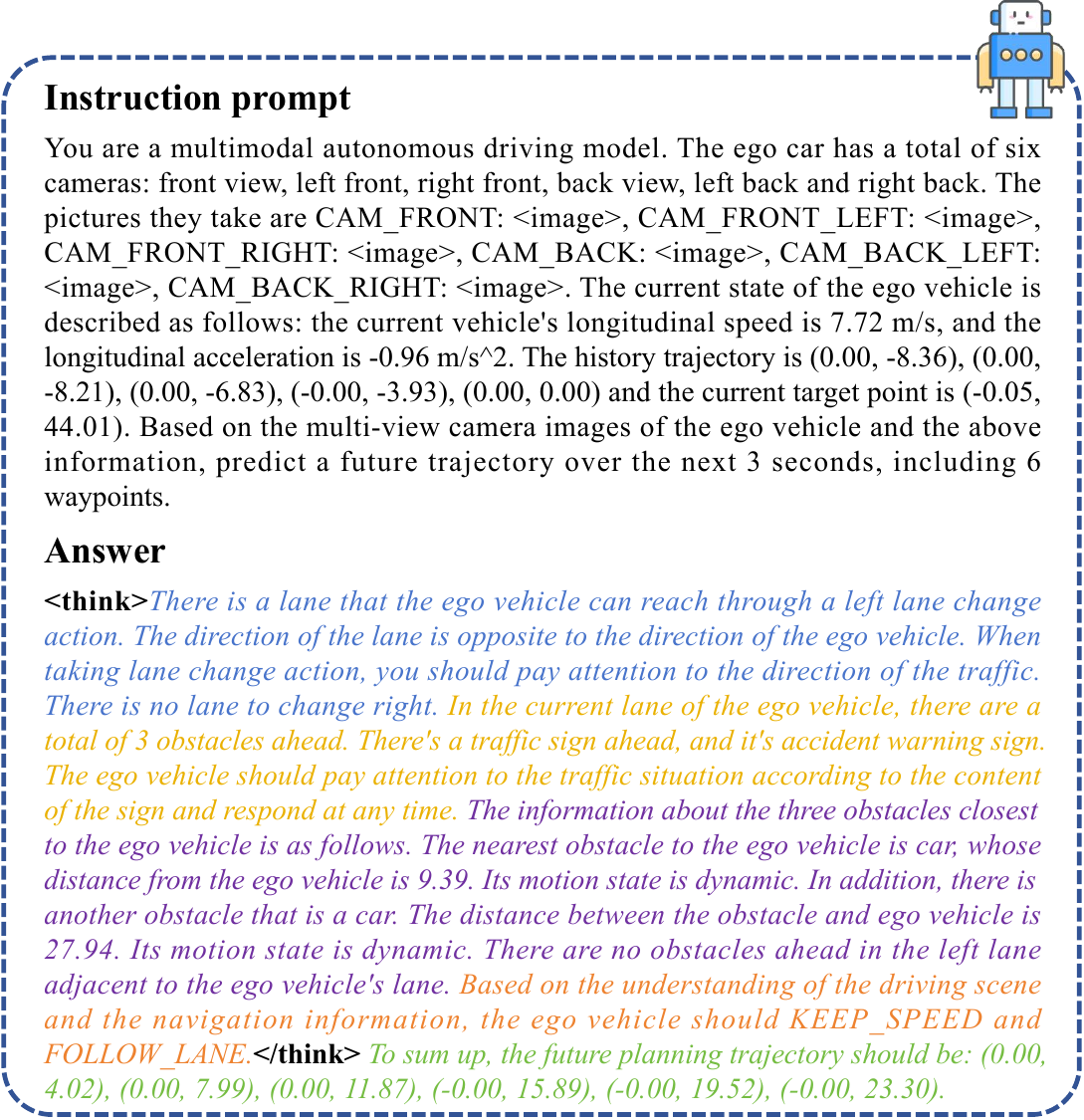}
    \caption{\small An Complete sample of the PDR dataset. }
    \label{fig: data_example}
\end{figure}

\subsection{Data Statistics}

\begin{figure}[t]
    \centering
    \begin{subfigure}[t]{0.8\textwidth}
        \centering
        \includegraphics[width=\textwidth]{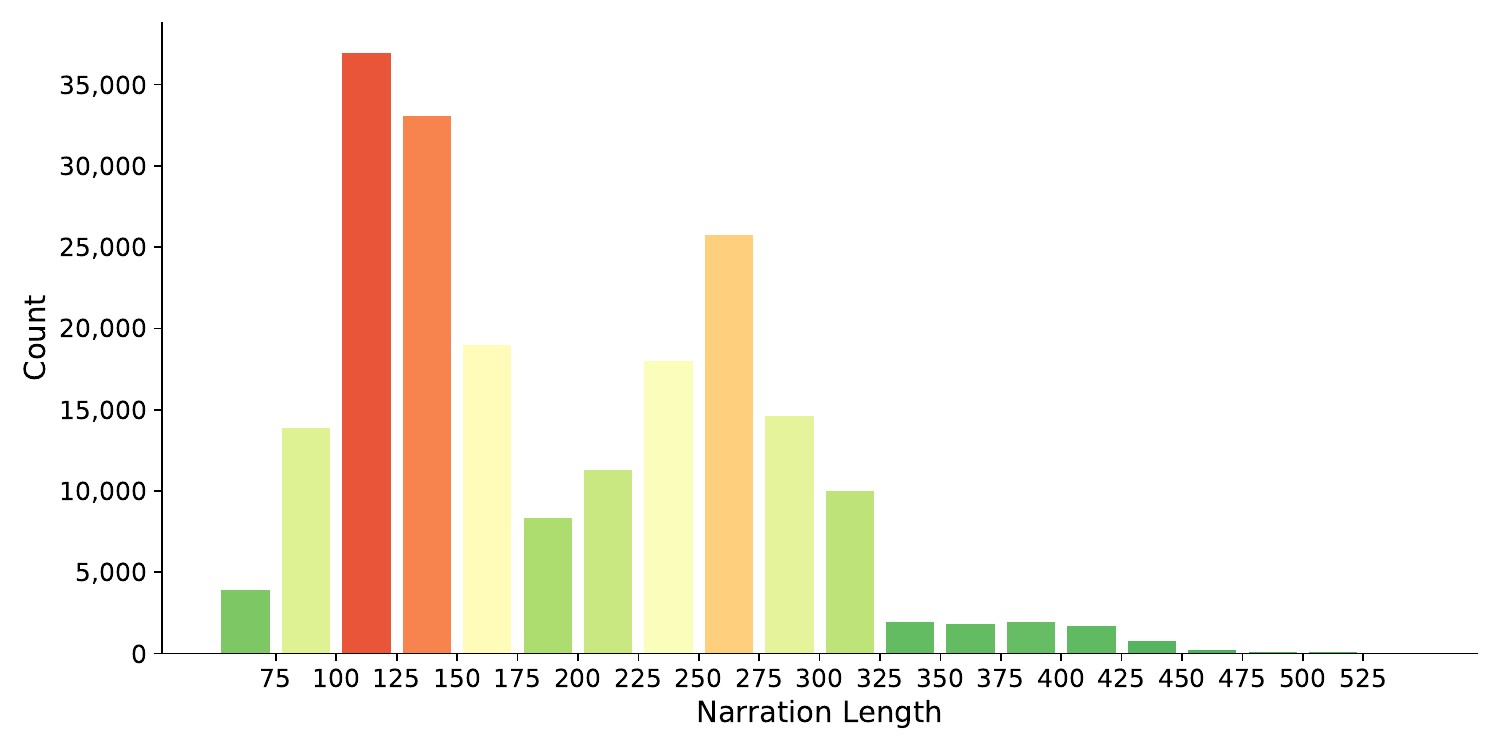}
        \caption{}
        \label{fig:case_b2d}
    \end{subfigure}
    \vspace{0.3cm}
    
    \begin{subfigure}[t]{0.8\textwidth}
        \centering
        \includegraphics[width=\textwidth]{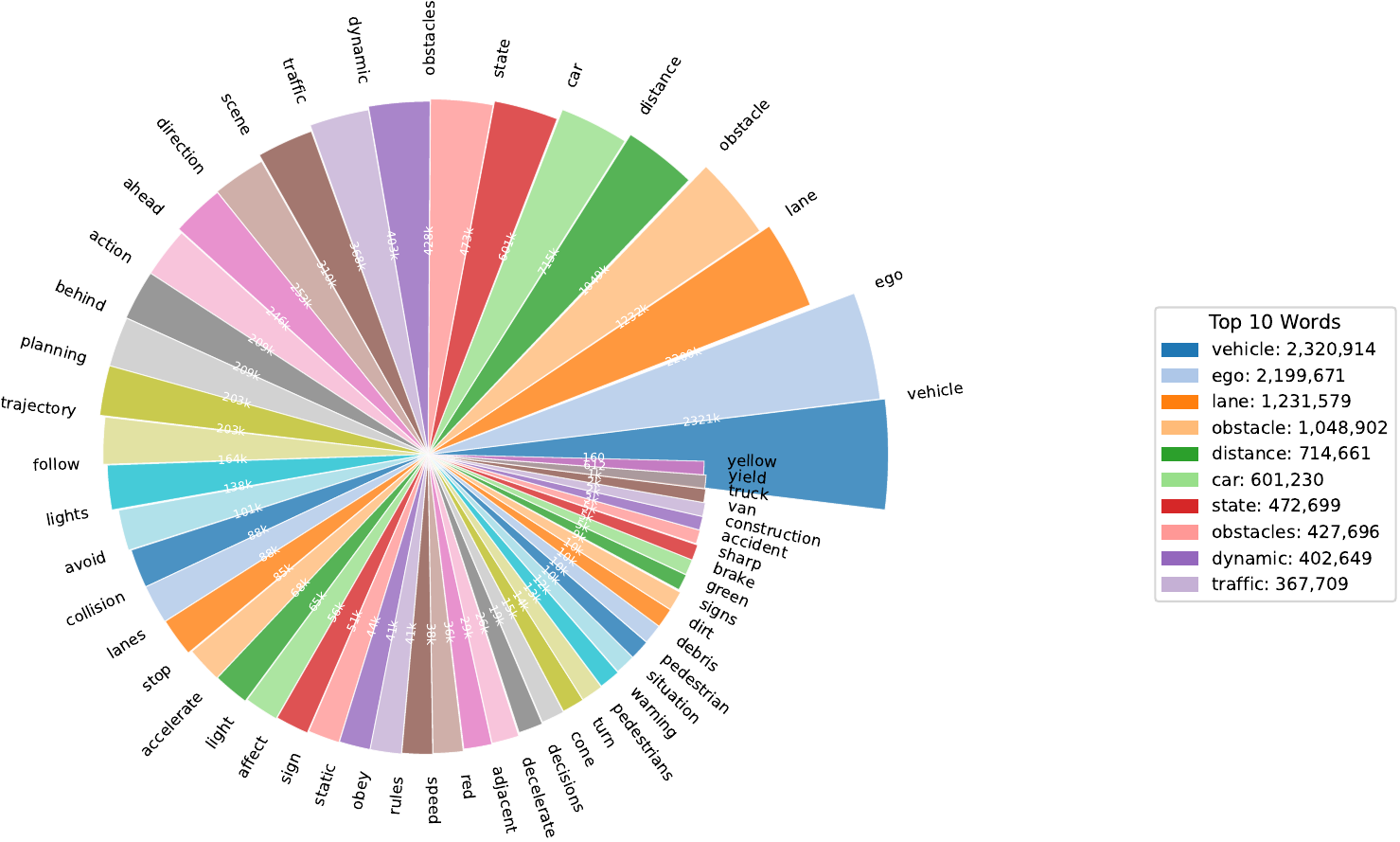}
        \caption{}
        \label{fig:case_dos}
    \end{subfigure}

    \caption{Data statistics of PDR. (a) Distribution of reasoning sentence length. (b) The key vocabulary chart of PDR. }
    \label{fig:data_statistics}
    \vspace{-10pt}
\end{figure}

Through our automated pipeline, we constructed a high-quality reasoning dataset containing 203,353 annotated instances, with an average of 195 words per image frame. Fig.~\ref{fig: data_example} shows an Complete sample of the PDR dataset. The dataset’s attributes are illustrated in Fig.~\ref{fig:data_statistics}. The narration length is up to 500 words, while the vocabulary chart indicates the planning-oriented information.

\section{Benchmarks}
\label{sup: benchmark}
\subsection{Bench2Drive}
Bench2Drive~\cite{jia2024bench2drive} designs a comprehensive and fair testing benchmark for evaluating E2E autonomous driving systems’ multiple abilities in a closed-loop manner.
Bench2Drive is an offline dataset collected using the RL-based agent in CARLA v2. We adopt the Base subset, which contains 230k frames extracted from 1,000 short clips. After filtering out the terminal frames of each scenario, a total of 203,353 samples are retained. These clips are uniformly distributed across 44 interactive driving scenarios, 23 weather conditions, and 12 towns.
The evaluation protocol requires E2E autonomous driving models to complete all 44 scenarios under various locations and weather conditions, resulting in 220 distinct routes. This setup enables a comprehensive and disentangled assessment of driving capabilities across diverse conditions.

Due to the high computational cost of full evaluation, we employ the Dev10 subset for ablation studies. It consists of 10 carefully selected routes from the official set, chosen to be both challenging and representative while maintaining low variance.

\subsection{DOS}
DOS is publicly released in ReasonNet~\cite{shao2023reasonnet}, a driving simulation benchmark in Carla consisting of diverse occlusion events, where E2E models have to reason about the global information of driving scenes.
DOS includes four types of challenging scenarios:
\begin{itemize}
    \item \textbf{Parked Cars (DOS01):} The ego vehicle drives along a straight lane with parked vehicles on both sides. Pedestrians initially appear on the visible sidewalk but may suddenly emerge from occluded regions between parked cars, creating abrupt crossing hazards.
    \item \textbf{Sudden Brake (DOS02):} The ego vehicle follows traffic in a straight lane. Pedestrians unexpectedly emerge from the sidewalk, prompting leading vehicles to brake abruptly. These pedestrians remain outside the ego vehicle's field of view, making anticipation challenging.
    \item \textbf{Left Turn (DOS03):} The ego vehicle attempts an unprotected left turn at an intersection. A large truck in the opposing lane intermittently occludes oncoming traffic, making it difficult to detect vehicles proceeding straight.
    \item \textbf{Red Light Infraction (DOS04):} The ego vehicle crosses an intersection behind several trucks. A vehicle running a red light approaches laterally but remains occluded by the trucks until the last moment, leading to a high-risk collision scenario.
\end{itemize}

 Each of the four scenarios in the DOS benchmark consists of 25 distinct cases with diverse road layouts and background traffic conditions, resulting in a total of 100 evaluation cases.

\newcommand{\parameters}
{{\footnotesize
\begin{tabular}{c|c|c}
\toprule
Parameters & Description & Value \\
\midrule
$L_v$ & the number of visual tokens per grid & 729 \\
$D_v$ & the visual embedding dimension & 1152 \\
$D_p$ & the textual embedding dimension & 1024 \\
$b_s$ & the training batch size & 16 \\
$l_r$ & the learning rate & 5e-5 \\
\bottomrule
\end{tabular}}}

\begin{table}[!ht]
\setlength\tabcolsep{3pt}
    \centering
    \caption{\small Other parameters. }
    \parameters
    \label{tab:parameters}
\end{table}
 
\section{Implementation Details}
\label{sup: implementation}
All experiments are conducted on 8 NVIDIA L20 GPUs (48 GB each). 
To accelerate the alignment of the vision-language space and gradually enhance the reasoning and planning capabilities of ReasonPlan, we adopt a two-stage training strategy. In each stage, the model inherits the weights from the previous stage and continues training.
The first stage consists of 1 epoch and completes in approximately 6 hours, while the second stage spans 3 epochs, requiring around 23 hours. The other parameters are as shown in Tab.~\ref{tab:parameters}.

\newcommand{\Ablationviews}
{{\footnotesize
\begin{tabular}{c|c|c|c}
\toprule
\textbf{Views} & RC$\uparrow$ & IS$\uparrow$ & DS$\uparrow$ \\
\midrule
Front-view & \textbf{89.29} & 0.65 & 57.83 \\
All-views & 87.35 & \textbf{0.66} & \textbf{58.95} \\
\bottomrule
\end{tabular}}}


\newcommand{\Ablationweight}
{{\footnotesize
\begin{tabular}{c|c|c|c}
\toprule
\textbf{Image Weights} & RC$\uparrow$ & IS$\uparrow$ & DS$\uparrow$ \\
\midrule
0 & 73.73 & \textbf{0.76} & 53.97 \\
0.25 & 80.12 & 0.68 & 56.96 \\
0.5 & 79.52 & 0.75 & \textbf{61.13} \\
1  & \textbf{89.29} & 0.65 & 57.83 \\
2  & 80.80 & 0.66 & 54.44 \\
\bottomrule
\end{tabular}}}

\noindent
\begin{minipage}{0.54\linewidth}
  \centering
  \captionof{table}{Ablation on NSP Views.}
  \renewcommand{\arraystretch}{0.9}
  \Ablationviews
  \label{tab:ablation-views}
\end{minipage}
\hfill
\begin{minipage}{0.48\linewidth}
  \centering
  \captionof{table}{Ablation on Image Weights}
  \renewcommand{\arraystretch}{0.9}
  \Ablationweight
  \label{tab:ablation-weights}
\end{minipage}

\section{More Ablation Experiments}
\label{sup: ablation}
\textbf{Front-view prediction performs on par with using all views.} For the next scene prediction (NSP) task, we conduct an ablation comparing models trained with only the front-view image versus those utilizing all camera views. As shown in Tab.~\ref{tab:ablation-views}, the results indicate that front-view-only prediction yields performance comparable to using the full view set. This suggests that the front-facing image captures the most semantically informative content for driving decisions, including cues critical for trajectory planning. To improve computational efficiency without compromising performance, we adopt the front view exclusively in all subsequent experiments.

\textbf{A synergy–conflict dynamic exists between NSP and textual reasoning.} As shown in Tab.~\ref{tab:ablation-weights}, when the image loss weight is too low, the supervisory signal from NSP becomes negligible, offering limited benefit to language-based reasoning. In contrast, overly high image weights lead the model to overfit to visual prediction, impairing its reasoning capability in the textual domain. Empirically, setting the image weight to 0.5 provides a balanced trade-off, enabling effective interaction between NSP and DeCoT and yielding the best overall performance.

\begin{figure}[t]
    \centering
    \begin{subfigure}[t]{1\textwidth}
        \centering
        \includegraphics[width=\textwidth]{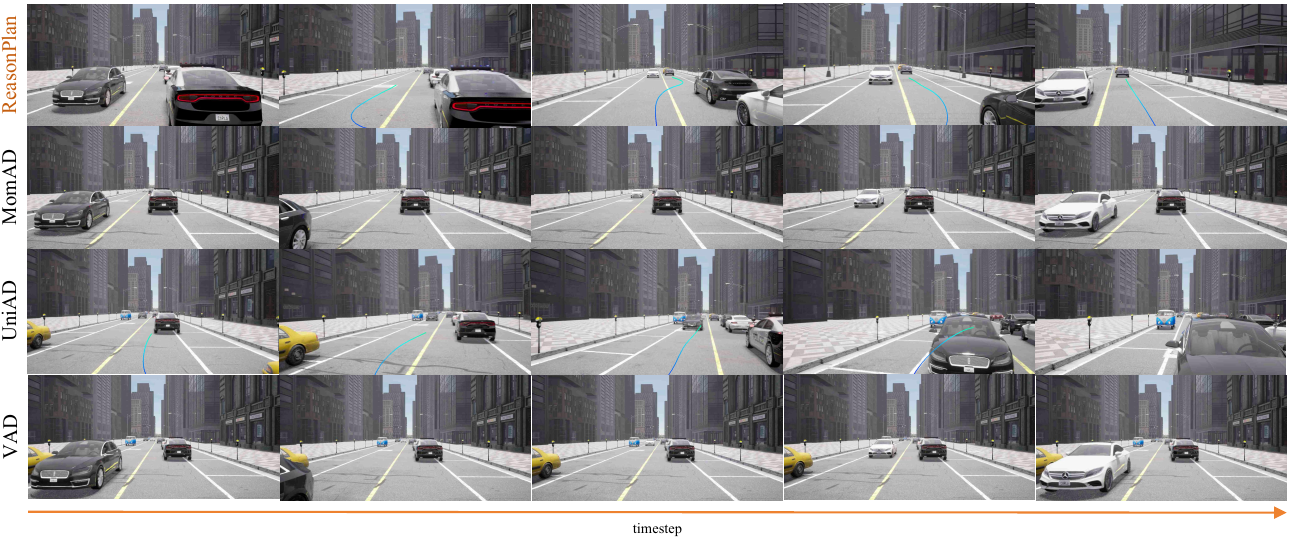}
        \caption{\small AccidentTwoWays in Bench2Drive. The autonomous vehicle must execute two consecutive lane changes while maintaining awareness of oncoming traffic. MomAD and VAD, due to their conservative policies, come to a halt in front of the obstructing vehicle; UniAD, on the other hand, takes a more aggressive approach, disregarding the oncoming traffic, resulting in a collision during the lane change. In contrast, ReasonPlan accurately perceives the relative positions of surrounding vehicles, avoiding two successive oncoming vehicles in a timely manner and successfully completing the lane change task. }
        \label{fig:case_1}
    \end{subfigure}
    \vspace{0.3cm}
    
    \begin{subfigure}[t]{1\textwidth}
        \centering
        \includegraphics[width=\textwidth]{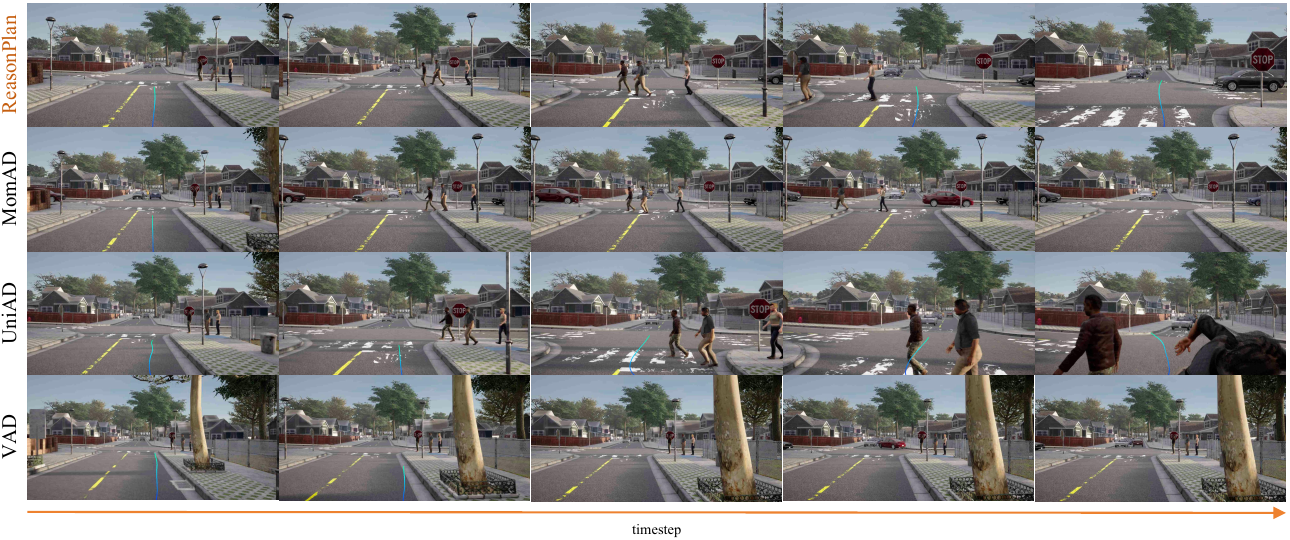}
        \caption{\small PedestrianCrossing in Bench2Drive. After avoiding the pedestrian, both MomAD and VAD fail to resume their motion, while UniAD cannot react to the pedestrian’s presence, leading to a collision. ReasonPlan, however, promptly halts upon detecting the pedestrian and successfully resumes operation once the pedestrian has cleared, completing the task efficient.}
        \label{fig:case_2}
    \end{subfigure}

    \caption{\small Additional Qualitative Comparison Cases for ReasonPlan.}
    \label{fig:good_case}
    \vspace{-10pt}
\end{figure}

\section{More Case Study}
\label{sup: case}
We present two additional representative comparison cases in Fig.~\ref{fig:good_case}. In contrast, existing methods generally exhibit delayed responses or policy failures in such high-risk scenarios, further highlighting the advantages of ReasonPlan in terms of decision-making reliability and safety.

\begin{figure}[t]
    \centering
    \begin{subfigure}[t]{1\textwidth}
        \centering
        \includegraphics[width=\textwidth]{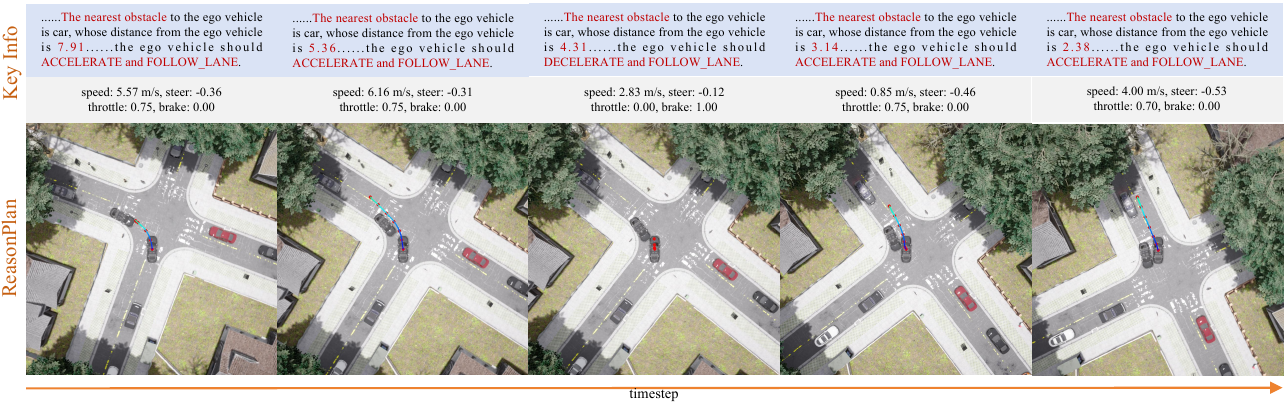}
        \caption{\small Although identifying proximal obstacles, ReasonPlan exhibited failure in obstacle avoidance. }
        \label{fig:bad case_1}
    \end{subfigure}
    \vspace{0.3cm}
    
    \begin{subfigure}[t]{1\textwidth}
        \centering
        \includegraphics[width=\textwidth]{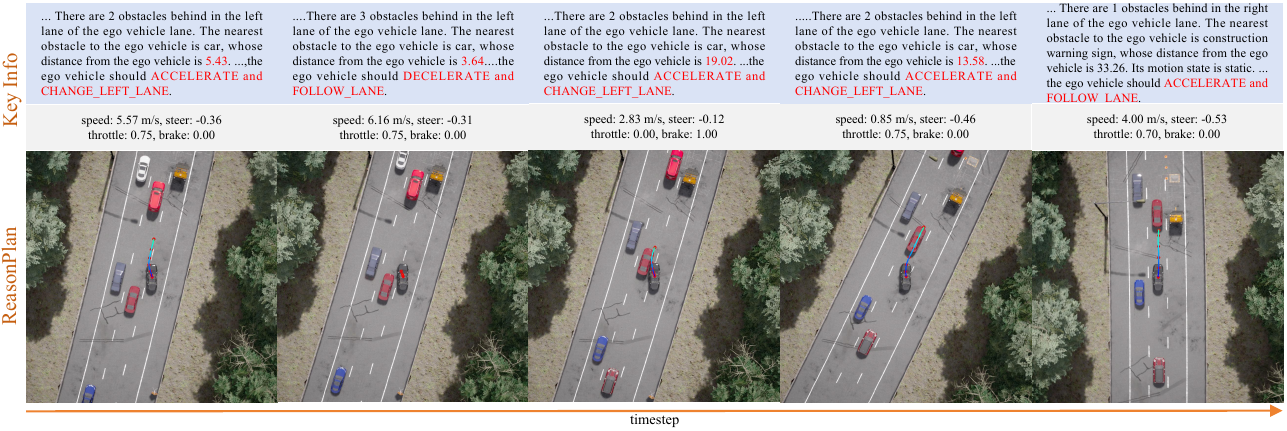}
        \caption{\small ReasonPlan accurately detected the approaching vehicle from behind but failed to complete the lane change maneuver within a short time frame, resulting in a collision.}
        \label{fig:bad case_2}
    \end{subfigure}

    \caption{\small Failed Cases for ReasonPlan.}
    \label{fig:bad_case}
    \vspace{-10pt}
\end{figure}

Although our fine-tuning framework enhances the model's reasoning capability in complex scenarios, failure modes still exist in some cases. As shown in FIg.~\ref{fig:bad_case}, ReasonPlan still encounters unavoidable collision scenarios. As noted in the Limitations section, the method relies solely on supervised fine-tuning with offline datasets and cannot incorporate feedback-driven learning. Reinforcement learning, as a trial-and-error learning paradigm, may present a promising direction to mitigate this constraint.

\end{document}